\documentclass[letterpaper, 10 pt, journal, twoside]{IEEEtran}
\usepackage{cite}
\usepackage{amsmath,amssymb,amsfonts}
\usepackage{algorithmic}
\usepackage{graphicx}
\usepackage{textcomp}
\usepackage{xcolor}
\usepackage{multirow}
\usepackage{makecell}
\usepackage{threeparttable}
\usepackage{array}
\usepackage{booktabs}
\usepackage{amsbsy}
\usepackage{bbm}
\usepackage{algorithm}
\usepackage{algorithmic}

\ifCLASSINFOpdf
\else
\fi
\begin{document}
%
\title{IFTD: Image Feature Triangle Descriptor for Loop Detection in Driving Scenes}
%
%
%

\author{Fengtian~Lang$^{1}$, Ruiye~Ming$^{1}$, Zikang~Yuan$^{2}$ and Xin~Yang$^{1*}$
\thanks{$^{1}$Fengtian~Lang, Ruiye~Ming and Xin~Yang$^{*}$ are with the Electronic Information and Communications, Huazhong University of Science and Technology, Wuhan, 430074, China. (* represents the corresponding author. E-mail: {\tt\small M202372913@hust.edu.cn; M202272555@hust.edu.cn; xinyang2014@hust.edu.cn})}
\thanks{$^{2}$Zikang~Yuan is with Institute of Artificial Intelligence, Huazhong University of Science and Technology, Wuhan, 430074, China. (E-mail: {\tt\small yzk2020@hust.edu.cn})}%
}
%
%

\markboth{IEEE Robotics and Automation Letters. Preprint Version. Accepted Month, Year}
{FirstAuthorSurname \MakeLowercase{\textit{et al.}}: ShortTitle}

%



\maketitle
\pagestyle{empty}
\thispagestyle{empty}
\begin{abstract}
	\par In this work, we propose a fast and robust Image Feature Triangle Descriptor (IFTD) based on the STD method, aimed at improving the efficiency and accuracy of place recognition in driving scenarios. We extract keypoints from BEV projection image of point cloud and construct these keypoints into triangle descriptors. By matching these feature triangles, we achieved precise place recognition and calculated the 4-DOF pose estimation between two keyframes. Furthermore, we employ image similarity inspection to perform the final place recognition. Experimental results on three public datasets demonstrate that our IFTD can achieve greater robustness and accuracy than state-of-the-art methods with low computational overhead.
\end{abstract}

%
\IEEEpeerreviewmaketitle

\section{Introduction}
\label{Introduction}

\par \IEEEPARstart{P}lace recognition, also known as loop detection, is the process of identifying and matching the current position of a robot with previously visited locations in a known environment. Loop detection is important for correcting odometry drift and ensuring global consistency. Due to the widespread use of visual sensors, many vision-based loop detection methods, such as DBoW \cite{galvez2012DBoW2}, have become popular. However, visual sensors are susceptible to challenges such as lighting conditions, viewpoint changes, and occlusions, making loop detection difficult. Compared to visual sensors, LiDAR is invariant to lighting and appearance changes and can directly capture the structural information of the surrounding environment, thus enabling more accurate loop detection.
\par LiDAR point based loop detection methods should possess the following three fundamental characteristics: (1) the descriptor for place recognition should possess rotational and translational invariance regardless of viewpoint changes. (2) Considering the complexity of 3D point clouds, the construction and retrieval of descriptors must be efficient to swiftly detect loops. (3) The similarity constraint should be sufficiently reliable to solve the relative pose between the current frame and the loop frame. Guided by the above fundamental requirements, Yuan et al. proposed the STD \cite{yuan2023std} algorithm, which directly extracts planar boundary feature points from 3D point clouds to construct feature triangles.The descriptors generated by this method possess invariance to rotation and translation, but the computational cost of extracting feature points and constructing feature triangles is relatively high. Furthermore, in complex environments such as forests, the method is difficult to identify stable planar edge feature points, limiting its applicability in such challenging scenarios.
\par To address this issue, we propose IFTD, an image feature triangle descriptor based on BEV images, suitable for driving scenarios. First, we perform height-based layering of the 3D point cloud, followed by BEV projection based on the layered heights. Next, we extract Shi-Tomasi feature points \cite{shi1994good} from the BEV images, which are invariant to rotation and translation, and construct feature triangles from these feature points. Subsequently, we determine whether the loop closure has occurred based on the similarity of these feature triangles and BEV images. .Experimental results on the NCLT \cite{carlevaris2016NCLT}, KITTI \cite{geiger2012KITTI}, and Mulran \cite{kim2020mulran} datasets demonstrate that IFTD shows excellent accuracy and robustness on most sequences compared to state-of-the-art methods.In addition, it is significantly better than STD in terms of speed and robustness.
\par Specifically, our contributions are as follows:
\begin{itemize}
	\item We designed a triangle descriptor based on BEV image features, which is more robust in complex environments and achieves faster computational efficiency.
	\item We proposed a rapid geometric verification method based on BEV projection images. which can quickly identify and match scene similarities and provide precise 4-DoF (x, y, z, and yaw) pose estimation.
	\item We have released the source code of our system to benefit the development of the community \footnote{https://github.com/EinsTian1/iftd}.
\end{itemize}
\par The rest of this paper is organized as follows: In Sec.\ref{Related Work}, we provide a brief overview of relevant literature. Sec.\ref{METHODOLOGY} elaborates on the proposed Image Feature Triangle Descriptor. Sec.\ref{Experiments} presents the experimental evaluation. Finally, we conclude the paper in Sec.\Ref{Conclusion}.

\section{Relad Work}
\label{Related Work}

\par Most 3D point based loop detection methods based are derived from vision based loop detection methods. In this section, we firstly provide a brief overview of vision based loop detection methods, and then detailedly discuss the 3D point based loop detection methods. Finally, we briefly discuss the learning based loop detection methods.
\subsection{Visual-based loop detection}
\par Visual-based loop detection methods have been widely applied in various visual SLAM frameworks \cite{mur2015orb, qin2018vins, lin2021r2LIVE, yuan2024srlivo,yuan2023sdv}. These methods typically extract visual feature descriptors from images and transform them into bag-of-words (BoW) vectors using DBoW \cite{galvez2012DBoW2} on a pre-trained visual vocabulary. Finally, loops are identified based on similarity scores between these BoW vectors. However, the detection performance of visual-based loop detection methods largely depends on the visual features of the environment. If there are significant changes in lighting conditions or viewpoints when the robot revisits a location, these methods may fail to provide stable and reliable detection results.

\subsection{LiDAR-based loop detection}
\par In recent years, many excellent LiDAR based LIOs have emerged, such as \cite{zhang2014loam, shan2020lio, xu2022fast2, yuan2023semi, yuan2022sr, chen2023direct}, which have an important demand for LiDAR based Loop Detection. For LiDAR point based loop detection methods, global descriptors are more robust to local noise and point cloud density, thus the LiDAR point based loop detection method tends to construct global descriptors. M2DP \cite{he2016m2dp} achieves loop detection by projecting the raw point cloud onto multiple 2D planes to generate signature vectors. Scan context \cite{kim2018scan} obtains the BEV projection of point clouds by recording the maximum height of the raw point cloud and proposes a fast loop detection method based on ring key registration. Scan context++ \cite{kim2021scan++} extends Scan context by adding enhanced descriptors to simulate lateral offset and reverse rotation of vehicles, making it suitable for urban environments. LiDAR-Iris \cite{wang2020lidar_iris} encodes point cloud height information to obtain the BEV projection of point clouds, and performs multiple LoG-Gabor filtering and thresholding operations to obtain binary signature images for each point cloud. However, this method adopts a brute-force search strategy, which results in excessive computational cost during loop searching and makes it difficult to apply in practical scenarios. Imaging-LiDAR \cite{shan2021imaging_lidar} converts the point cloud obtained by multi-line LiDAR into a 2D image based on depth information, and uses a visual method based on DBoW for loop detection. Since this method requires multi-line LiDAR, it is difficult to apply in practical scenarios. Contour Context \cite{jiang2023contour} obtains different height-layered BEV projections by height layering and extracts contour information by detecting the connectivity of different layers, thereby constructing constellation descriptors. This algorithm utilizes a two-step similarity test of discrete constellation consistency verification and continuous density L2 optimization to achieve loop detection. STD \cite{yuan2023std} extracts planes from the original point cloud, detects feature points on the edges of these planes, and then uses these feature points to construct triangle descriptors for loop detection. Moreover, this method uses hash voxels to store descriptors, achieving faster descriptor retrieval speed compared to traditional kd-tree methods.
\subsection{Learning-based loop detection}
\par Deep learning technology has shown great potential in the field of loop detection, with its advantage of capturing structural information of point cloud data from a global perspective. OverlapNet \cite{chen2021overlapnet} utilizes deep learning to extract different structural information from point cloud and evaluates the degree of overlap between these point cloud information to determine whether there is a loop. OverlapTransformer \cite{ma2022overlaptransformer}, introduces Transformer attention mechanisms on the basis of OverlapNet, significantly enhancing the algorithm's invariance to rotational changes. LCD-Net \cite{cattaneo2022lcdnet} combines the ability of DNNs to extract unique features from point clouds with feature matching algorithms extracted from transport theory, achieving 6-DoF pose estimation.

\section{METHODOLOGY}
\label{METHODOLOGY}

\par
\begin{figure}[t]
	\begin{center}
		\includegraphics[scale=0.50]{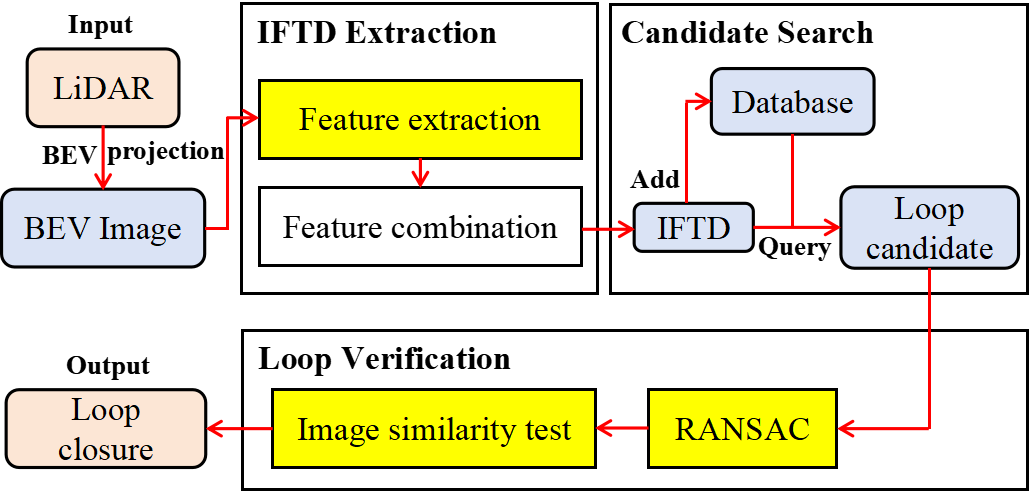}
		\caption{Framework of our loop detection method. The yellow part is the core contribution we proposed.}
		\label{fig1}
	\end{center}
\end{figure}

 In this section, we will introduce how to construct image feature triangle descriptors and improve the method of STD. The diagram of our method is shown in Fig.\ref{fig1}. The function of the IFTD extraction module is to extract feature points from LiDAR BEV projection images and then constructs triangle descriptors. The function of the candidate search module is to retrieve the top 50 frames from the database that match the current frame in terms of triangle descriptors In the loop verification module, the RANSAC algorithm uses SVD decomposition to select matching candidate frames, while image similarity testing determines the loop closure by comparing the current frame with the matching candidate frames.

\subsection{BEV Projection Image Acquisition}
\label{BEV Projection Image Acquisition}
\begin{figure}[b]
	\begin{center}
		\includegraphics[scale=0.50]{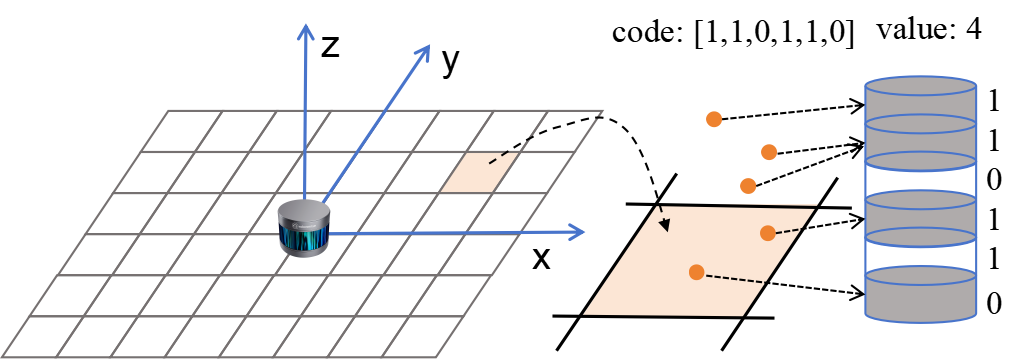}
		\caption{Illustration of Encoding the height information of 3D point clouds to obtain a Bird's Eye View (BEV) projection image.}
		\label{fig2}
	\end{center}
\end{figure}

\par
\begin{figure}
	\begin{center}
		\includegraphics[scale=0.60]{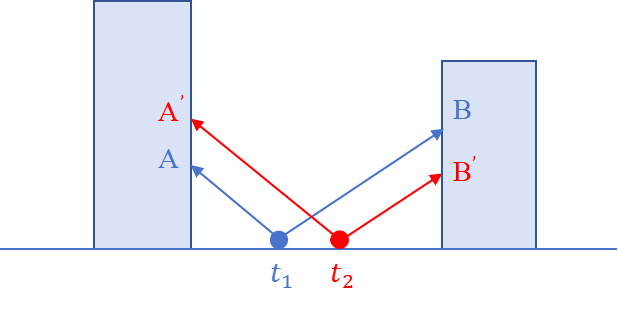}
		\caption{The impact of translation on maximum height. From $t_1$ to $t_2$, the vehicle occurred large lateral movement. At time $t_1$, the highest point cloud data collected is labeled as points $A$ and $B$, while at time $t_2$, the highest point cloud is labeled as points $A'$ and $B'$. By comparing these points, we can clearly observe the effect of lateral translation of the vehicle on the maximum height in the point cloud measurements.}
		\label{fig3}
	\end{center}
\end{figure}

 When processing a point cloud, the first step is to project it onto a Bird's Eye View (BEV). We define a square region of k $\times$ k square meters as the effective sensing area, with the LiDAR positioned at the center of this area. Next, we divide the effective sensing area into N $\times$ N bins, corresponding to the resolution of the BEV image. Subsequently, based on the (x, y) coordinates of the point within the effective sensing area, we calculate the bin in which the point lies and assign a value to the bin. In Scan Context++ \cite{kim2021scan++}, the maximum height of the point cloud within each bin is used as the value for that bin. However, when the vehicle occurs large lateral offset, the maximum height of each bin will change accordingly. as shown in Fig.\ref{fig2}, which may affect the accuracy of loop detection. To mitigate the impact of offset on point cloud height, we adopt a height encoding approach to record bin values. Specifically, we further divide the point cloud within each bin into multiple different height layers. If there are LiDAR points within a particular height layer, the layer is marked as 1. Therefore, we can obtain a height encoded binary sequence of point clouds in each bin. Finally, we take the total number of 1 in this binary sequence as the value of the bin, as shown in Fig.\ref{fig3}. The advantage of this method lies in its ability to reflect the vertical distribution of the point cloud within each bin. A higher bin value indicates a wider vertical distribution of the point cloud within the bin, making it a potential feature point. Through this approach, we effectively avoid the adverse effects caused by outliers with higher heights on feature point detection, thereby improving the stability and accuracy of loop detection.

\subsection{Image Feature Triangle Descriptor Extractiontle}
\label{Image Feature Triangle Descriptor Extraction}
\par
\begin{figure}
	\begin{center}
		\includegraphics[scale=0.55]{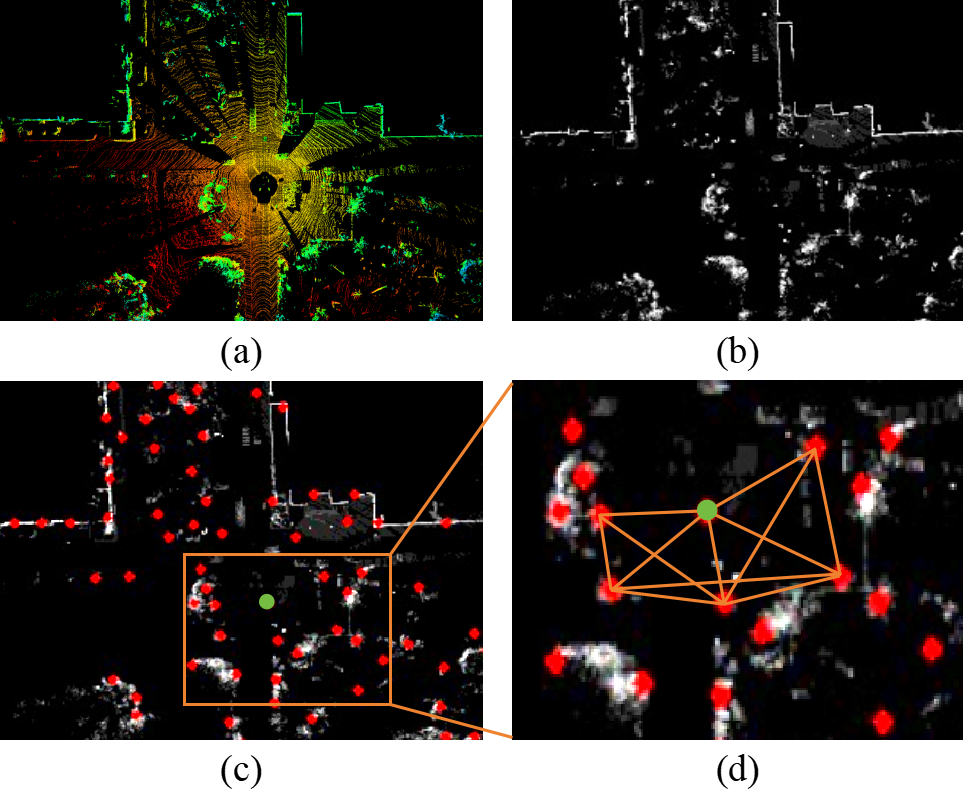}
		\caption{Illustration of Image Feature Triangle Descriptor. Fig.a represents the original 3D point cloud. Fig.b represents the BEV projection image of the point cloud. In Fig.c, the red dots represent the feature points extracted from the BEV image using the Shi-Tomasi method. Fig.d shows the feature triangles constructed with the green point at the center and its neighboring points.}
		\label{fig4}
	\end{center}
\end{figure}

 Through the above-mentioned steps, we successfully generated the corresponding Bird's Eye View (BEV) projection image(as shown in Fig.\ref{fig4} (b)) from the raw point cloud (as shown in Fig.\ref{fig4} (a)). This BEV projection image is typically sparse but clearly reflects the geometric structure distribution of the point cloud. Given the characteristics of the BEV image, we chose the Shi-Tomasi method for reliably extracting feature points. The Shi-Tomasi method is a widely recognized feature detection algorithm known for its stability and reliability, and can successfully extract feature points from the BEV image (as shown in Fig. 3 (c)). These feature points highlight geometric characteristics of the point cloud in the longitudinal distribution. In this manner, we can more precisely ascertain and align environmental landmarks, thereby enhancing the precision and robustness of loop detection.
\par We utilize the feature points extracted from the BEV image to construct a k-D tree and search for the 15 nearest neighbors for each point. As shown in Fig.\ref{fig4} (d), we use these neighboring points to construct feature triangle descriptors. To eliminate redundant feature triangles resulting from repeated construction between neighboring points, we remove triangles with the same vertices. Additionally, to ensure that extreme triangles (which can lead to mismatches) are not created due to collinear points, we apply an angle filter to the constructed feature triangles. This step ensures that all triangle angles remain within the range of 5° to 175°, excluding triangles with angles that are too small or too large as they might not provide effective geometric information. Finally, our feature triangle descriptors include the pixel values of the three vertices forming the triangle, the lengths of the three sides arranged in ascending order and the frame number corresponding to the descriptor. The data structure of the IFTD descriptor is shown in Data structure \ref{alg: IFTD Descriptor}.

\floatname{algorithm}{Data structure}
\begin{algorithm}[hb]
	\setcounter{algorithm}{0}
	\caption{IFTD Descriptor}
	\label{alg: IFTD Descriptor}
	\begin{algorithmic}
		\STATE $\textbf{struct}$ IFTD$\_$Descriptor $\{$
		\STATE \qquad \textbf{Eigen::Vector3d} Vertice$\_$A;
		\STATE \qquad \textbf{Eigen::Vector3d} Vertice$\_$B;
		\STATE \qquad \textbf{Eigen::Vector3d} Vertice$\_$C;
		\STATE \qquad \textbf{Eigen::Vector3d} Side$\_$length;
		\STATE \qquad \textbf{int} frame$\_$num;
		\STATE $\}$;
	\end{algorithmic}
\end{algorithm}

\label{Loop Verification}
\renewcommand{\algorithmicrequire}{ \textbf{Input:}} 
\renewcommand{\algorithmicensure}{ \textbf{Output:}} 

\floatname{algorithm}{Algorithm}
\begin{algorithm}[ht]
	\setcounter{algorithm}{0}
	\caption{Loop Verification}
	\label{alg: Loop Verification}
	\begin{algorithmic}[1] 
		\REQUIRE ~~\\ 
		Loop frame candidate $L$ and its cluster of triangle descriptor matches $S$;\\
		current frame $P_n$;\\
		\ENSURE ~~\\ 
		relative pose transformation $T$; \\
		geometric verification score $F$;
		\FOR{each $S_i \in S$}
		\STATE Solve for the pose transformation $T_i$ of the triangle match pair $S_i$ using SVD
		\FOR{each $S_j \in S$}
		\STATE Initialize $Dnum_i = 0$ and $Vnum_i = 0$
		\STATE Calculate the distances $dis_1$,$dis_2$,$dis_3$ of the three vertices of $S_j$ after transformation by $T_i$.
		\IF{$dis_1 < th(dis)$ and $dis_2 < th(dis)$ and $dis_3 < th(dis)$}
		\STATE $Dnum_i++$
		\STATE Calculate the number of non-overlapping triangle vertices $Vnum_i$
		\ENDIF
		\STATE Add $Dnum_i$ into the set $Dnum$, $Vnum_i$ into the set $Vnum$
		\ENDFOR
		\ENDFOR
		\STATE Find the maximum value $Dnum_{max}$ in the set of $Dnum$, its corresponding vertice count $Vnum_{max}$ and pose transformation $T_{max}$
		\IF{$Dnum_{max} > th({num}_{triangle})$ and $Vnum_{max} > th({num}_{vertice})$}
		\STATE Align BEV images of current frame $P_c$ and loop frame candidate $L$ according to $T_{max}$
		\STATE Downsample $P_c$ and $L$ to 20 $\times$ 20 resolution to get ${P_c^{'}}$ and ${L^{'}}$
		\STATE Compute the pixel-wise differences between adjacent pixels of ${P_c^{\prime}}$ and ${L^{\prime}}$ to obtain hash values $H_{P_c}$ and $H_L$
		\STATE Calculate the cosine similarity $F = cosine_similarity(H_{P_c}), H_L$
		\IF{$F > th(sim)$}
		\RETURN $T_{max}$ and $F$
		\ENDIF
		\ENDIF
	\end{algorithmic}
\end{algorithm}

\subsection{Candidate Search}
\label{Candidate Search}
\par We employ the same hash voxel-based loop candidate search algorithm as STD. Due to the rotation and translation invariance of the side length of a triangle, we use these three lengths as hash keys in the descriptor, along with recording the frame number corresponding to each descriptor. Descriptors with similar side length properties share the same hash key and are therefore stored in the same voxel container. When given a descriptor, we first locate the corresponding voxel container based on the hash value of the side lengths. Then, we iterate through all descriptors stored in that container and record their corresponding frame numbers. Given that a keyframe can extract thousands of feature descriptors, we employ a voting mechanism to determine the overlap count of triangle descriptors between historical frames and the current frame. After processing all descriptors in the current frame, we select the top 50 historical frames based on the voting results and save these matching descriptors. These saved descriptors will be used in subsequent fine-matching steps.

\subsection{Loop Verification}

\par When given a loop candidate, we conduct a fine-matching process to eliminate false detections caused by incorrect descriptor matching and calculate the relative pose transformation between the current frame and the true loop frame. Similar to STD, we utilize the correspondence between matched vertices and calculate the relative pose transformation $\mathbf{T} = \{ \mathbf{R}, \mathbf{t} \} $ between these two keyframes using Singular Value Decomposition (SVD), while employing the RANSAC algorithm to find the relative pose transformation that maximizes the number of correctly matched descriptors.

\par What differs from STD is that we don't use the method of detecting plane overlap to validate the similarity between two key frames in geometric verification. Instead, we adopt an image similarity-based detection method based on BEV images. Initially, we downsample the N $\times$ N resolution image to a 20 $\times$ 20 resolution. Then, by comparing the pixel-wise differences between the current pixel and its right adjacent pixel, generating a 20 $\times$ 19 matrix of image differences. Finally, we calculate the cosine similarity of the image difference matrices of the two images. Only when the image similarity exceeds a predefined threshold and is the maximum among all candidate frames, do we consider that frame as the loop frame corresponding to the current frame.The detailed algorithm is shown in Algorithm \ref{alg: Loop Verification}.

\begin{table}[h]
	\setlength{\tabcolsep}{10pt}
	\renewcommand{\arraystretch}{1.4}
	\begin{center}
		\caption{Datasets of All Sequences for Evaluation}
		\label{table1}
		\scalebox{1.0}{
			\begin{tabular}{c|c|c|c}
				\hline
				Dataset                 & Sequence    & \begin{tabular}[c]{@{}c@{}}Duration\\ (min:sec)\end{tabular} & \begin{tabular}[c]{@{}c@{}}Distance\\ (km)\end{tabular} \\ \hline
				\multirow{4}{*}{KITTI}  & KITTI00     & 7:50                                                         & 3.72                                                    \\
				                        & KITTI02     & 8:03                                                         & 5.07                                                    \\
				                        & KITTI05     & 4:47                                                         & 2.21                                                    \\
				                        & KITTI08     & 7:02                                                         & 3.22                                                    \\ \hline
				\multirow{9}{*}{Mulran} & DCC01       & 9:00                                                         & 4.91                                                    \\
				                        & DCC02       & 12:27                                                        & 4.27                                                    \\
				                        & DCC03       & 12:20                                                        & 5.42                                                    \\
				                        & KAIST01     & 13:33                                                        & 6.12                                                    \\
				                        & KAIST02     & 14:46                                                        & 5.96                                                    \\
				                        & KAIST03     & 14:16                                                        & 6.25                                                    \\
				                        & Riverside01 & 9:05                                                         & 6.43                                                    \\
				                        & RiverSide02 & 13:29                                                        & 6.61                                                    \\
				                        & RiverSide03 & 17:18                                                        & 7.25                                                    \\
				\hline
				\multirow{3}{*}{NCLT}   & 2012-01-08  & 93:08                                                        & 6.50                                                    \\
				                        & 2012-02-02  & 97:33                                                        & 6.32                                                    \\
				                        & 2012-05-26  & 88:15                                                        & 6.48                                                    \\
				\hline
			\end{tabular}}
	\end{center}
\end{table}

\section{Experiments}
\label{Experiments}

\begin{table*}[h]
	\setlength{\tabcolsep}{7pt}
	\renewcommand{\arraystretch}{1.5}
	\begin{center}
		\caption{Precision, Recall, F1-score of IFTD, STD and Contour Context on KITTI, Mulran and NCLT datasets.}
		\label{table2}
		\begin{tabular}{c|c|ccc|ccc|ccc}
			\hline
			\multirow{2}{*}{Dataset} & \multirow{2}{*}{Sequence} & \multicolumn{3}{c|}{IFTD} & \multicolumn{3}{c|}{STD} & \multicolumn{3}{c}{Contour context}                                                                               \\
			\cline{3-11}
			                         &                   & Precision        & Recall       & F1-score                & Precision & Recall & F1-score        & Precision & Recall & F1-score        \\
			\hline
			\multirow{4}{*}{KITTI}   & KITTI00           & 0.994            & 0.903        & $\textbf{0.947}$         & 0.976     & 0.827  & 0.896           & 0.969     & 0.787  & 0.868           \\
			                         & KITTI02           & 0.816            & 0.519        & 0.635                   & 0.720     & 0.701  & 0.711           & 0.982     & 0.714  & $\textbf{0.827}$ \\
			                         & KITTI05           & 1.00             & 0.936        & $\textbf{0.967}$         & 0.981     & 0.848  & 0.910           & 1.00      & 0.696  & 0.821           \\
			                         & KITTI08           & 0.924            & 0.678        & 0.782                   & 0.922     & 0.789  & $\textbf{0.850}$ & 1.00      & 0.611  & 0.759           \\
			\hline
			\multirow{9}{*}{Mulran}  & DCC01             & 0.937            & 0.962        & $\textbf{0.949}$         & 0.986     & 0.120  & 0.214           & 0.988     & 0.724  & 0.836           \\
			                         & DCC02             & 0.949            & 0.929        & $\textbf{0.939}$         & 0.992     & 0.309  & 0.471           & 0.975     & 0.608  & 0.749           \\
			                         & DCC03             & 0.967            & 0.959        & $\textbf{0.963}$         & 0.951     & 0.088  & 0.162           & 0.984     & 0.496  & 0.660           \\
			                         & KAIST01           & 0.994            & 0.829        & $\textbf{0.904}$         & 0.962     & 0.250  & 0.396           & 1.00      & 0.632  & 0.775           \\
			                         & KAIST02           & 0.992            & 0.867        & $\textbf{0.926}$         & 0.994     & 0.194  & 0.324           & 0.995     & 0.839  & 0.910           \\
			                         & KAIST03           & 0.991            & 0.951        & 0.971                   & 0.987     & 0.359  & 0.526           & 0.999     & 0.974  & $\textbf{0.986}$ \\
			                         & Riverside01       & 0.943            & 0.793        & 0.862                   & 0.447     & 0.091  & 0.152           & 0.910     & 0.881  & $\textbf{0.895}$ \\
			                         & RiverSide02       & 0.949            & 0.825        & 0.883                   & 0.726     & 0.167  & 0.272           & 0.972     & 0.826  & $\textbf{0.893}$ \\
			                         & RiverSide03       & 0.913            & 0.823        & 0.866                   & 0.773     & 0.245  & 0.372           & 0.959     & 0.920  & $\textbf{0.939}$ \\
			\hline
			\multirow{3}{*}{NCLT}    & 2012-01-08        & 0.937            & 0.763        & $\textbf{0.841}$         & 0.962     & 0.200  & 0.331           & 0.956     & 0.50   & 0.656           \\
			                         & 2012-02-02        & 0.961            & 0.808        & $\textbf{0.878}$         & 0.983     & 0.302  & 0.462           & 0.993     & 0.622  & 0.765           \\
			                         & 2012-05-26        & 0.969            & 0.769        & $\textbf{0.857}$         & 0.954     & 0.161  & 0.275           & 0.984     & 0.671  & 0.798           \\
			\hline
		\end{tabular}
	\end{center}
\end{table*}

\par We evaluated our method on the public datasets KITTI \cite{geiger2012KITTI}, Mulran \cite{kim2020mulran} and NCLT \cite{carlevaris2016NCLT}. KITTI dataset, which is captured by a 64-line Velodyne LiDAR, includes 11 sequences, with sequence 00, 02, 05, and 08 include loop closures. Hence, we select these four sequences for evaluation. Mulran is a dataset specifically designed to support place recognition evaluation, containing a large number of loops. This dataset contains a 64-line LiDAR scans (Ouster OS1-64). We selected KAIST01-03, DCC01-03 and Riverside01-03 for evaluation. KAIST01-03 and DCC01-03 sequences are collected from campus and urban environments with few dynamic objects. RiverSide01-03 sequence is collected along the road by the river. This sequence has many similar non-structured objects in the surrounding environment, such as trees on the roadside. Due to the sparse and similar geometric structures in the sequence, loop detection is more challenging. NCLT dataset is collected by a 32-line Velodyne LiDAR at the North Campus of the University of Michigan. All sequences of NCLT dataset were collected under the same scenario, and each sequence has an extensive duration of data collection. Thus, we selected three representative sequences (e.g., 2012-01-08, 2012-02-02, 2012-05-26) for evaluation.The detailed information of the datasets is shown in Table \ref{table1}.

\par To evaluate the effectiveness of loop detection in real-world scenarios, this experiment adopts a diverse approach to pose input. Given the lack of IMU data in the KITTI dataset, we use Ground Truth data to provide precise pose information. For the Mulran and NCLT datasets, we employ pose data generated by the Fast-LIO2 \cite{xu2022fast2} algorithm. This experimental design aims to test and verify the practical performance of various loop detection methods in correcting cumulative odometry errors when relying solely on odometry data as pose input. All experiments were conducted on the same system, using an Intel i7-12700 @ 4.9 GHz with 64GB of RAM.

\begin{table}[h]
	\setlength{\tabcolsep}{4pt}
	\renewcommand{\arraystretch}{1.5}
	\begin{center}
		\caption{Runtime Comparison of IFTD and STD on KITTI, Mulran and NCLT datasets.}
		\label{table3}
		\scalebox{1.0}{
			\begin{tabular}{c|ccc|ccc}
				\hline
				\multirow{2}{*}{Sequence} & \multicolumn{3}{c|}{IFTD} & \multicolumn{3}{c}{STD}                                      \\
				\cline{2-7}
				                          & extraction                & query                   & total & extraction & query & total \\
				\hline
				KITTI00                   & 4.46                      & 6.33                    & $\textbf{10.92}$ & 12.06      & 14.96 & 27.06 \\
				KITTI02                   & 3.46                      & 6.20                    & $\textbf{9.75}$  & 12.50      & 12.64 & 25.17 \\
				KITTI05                   & 4.64                      & 5.87                    & $\textbf{10.73}$ & 14.06      & 11.88 & 25.98 \\
				KITTI08                   & 5.26                      & 10.61                   & $\textbf{16.27}$ & 15.23      & 13.82 & 29.11 \\
				\hline
				DCC01                     & 4.91                      & 5.92                    & $\textbf{10.88}$ & 14.59      & 17.29 & 31.89 \\
				DCC02                     & 4.74                      & 21.61                   & $\textbf{26.38}$ & 14.98      & 23.36 & 38.36 \\
				DCC03                     & 5.10                      & 8.09                    & $\textbf{13.24}$ & 16.11      & 22.22 & 38.36 \\
				KAIST01                   & 5.48                      & 2.91                    & $\textbf{8.48}$  & 13.50      & 17.00 & 30.52 \\
				KAIST02                   & 5.16                      & 2.85                    & $\textbf{8.07}$  & 13.72      & 19.04 & 32.78 \\
				KAIST03                   & 4.46                      & 3.56                    & $\textbf{8.06}$  & 15.05      & 19.17 & 34.24 \\
				Riverside01               & 4.55                      & 2.35                    & $\textbf{6.96}$  & 9.53       & 9.29  & 18.82 \\
				RiverSide02               & 4.02                      & 3.31                    & $\textbf{7.38}$  & 10.49      & 13.04 & 23.55 \\
				RiverSide03               & 3.85                      & 6.29                    & $\textbf{10.1}$7 & 9.64       & 16.72 & 26.37 \\
				\hline
				2012-01-08                & 4.09                      & 8.05                    & $\textbf{12.21}$ & 13.44      & 35.61 & 49.06 \\
				2012-02-02                & 3.91                      & 7.87                    & $\textbf{11.83}$ & 13.74      & 39.31 & 53.07 \\
				2012-05-26                & 4.09                      & 6.80                    & $\textbf{10.93}$ & 13.47      & 31.74 & 45.23 \\
				\hline
			\end{tabular}}
	\end{center}
\end{table}

\subsection{Precision Recall Comparison with State-of-the-Arts}
\label{Precision Recall Comparison with State-of-the-Arts}

\begin{figure*}[h]
	\begin{center}
		\includegraphics[scale=0.24]{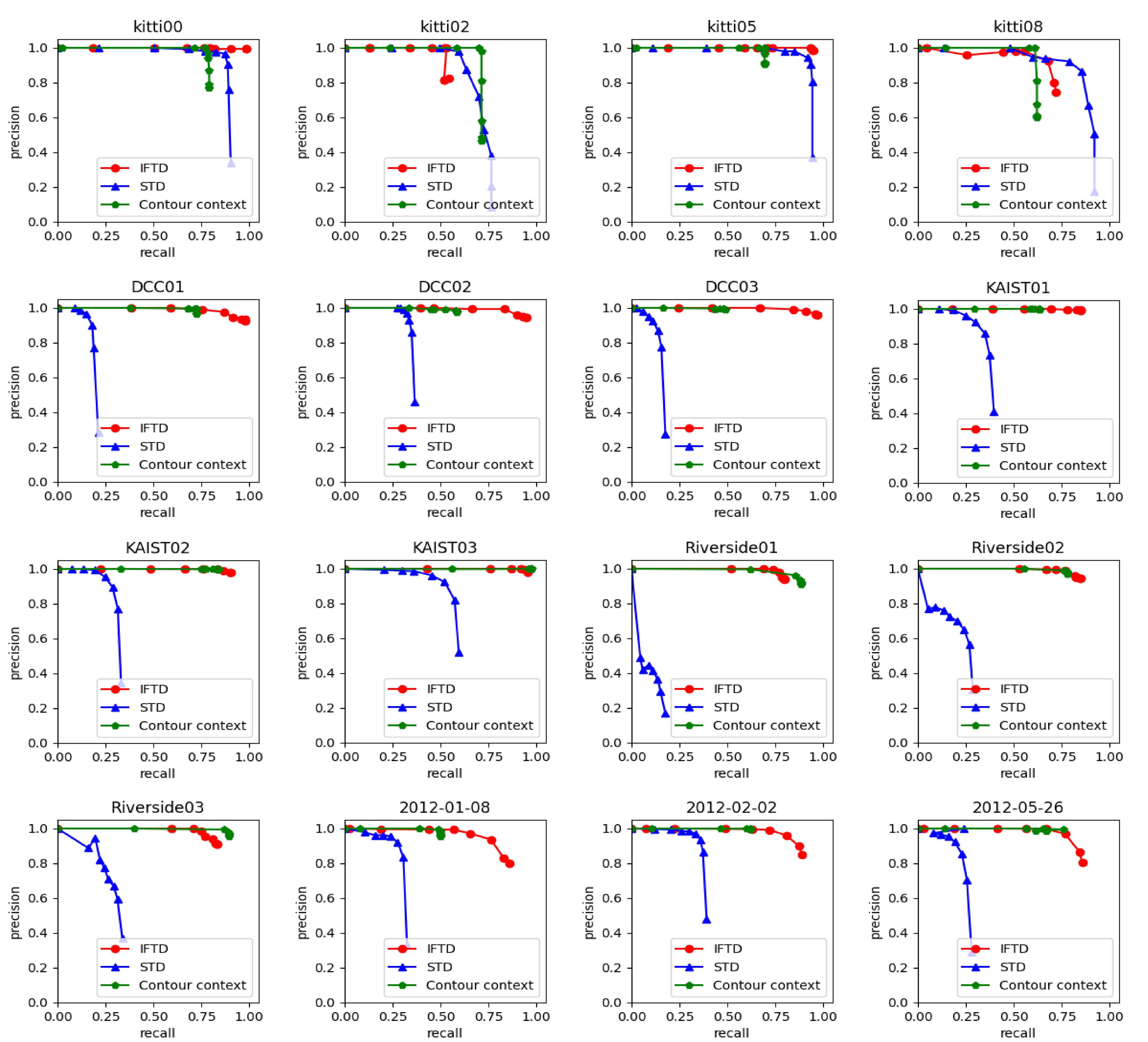}
		\caption{Precision-Recall curves of IFTD, STD, and Contour Context on KITTI, Mulran, and NCLT datasets.The red curve represents IFTD, the blue curve represents STD, and the green curve represents Contour Context.}
		\label{fig5}
	\end{center}
\end{figure*}

\par We compared our method with two other global descriptors: STD \cite{yuan2023std} and Contour Context \cite{jiang2023contour}. For both STD and Contour Context, we used their released code for experimental evaluation. When processing datasets of different lengths, we adopted different strategies to accumulate keyframes. For shorter datasets, such as KITTI and Mulran, we accumulated one keyframe every 5 frames. For longer datasets, like NCLT, we accumulated one keyframe every 10 frames. In our evaluation criteria, a detection is considered successful if the ground truth pose distance between the query keyframe and the matched keyframe is less than 15 meters.

\par Results in Table \ref{table2} demonstrate that our method achieves higher precision and recall results than STD and Contour Context on most sequences. To draw the precision-recall curves, we adjusted the threshold $\delta $ used in the geometric verification stage for several methods, thereby obtaining multiple pairs of precision and recall data at different threshold levels (as shown in Fig.\ref{fig5}). The results indicate that our proposed method outperforms the other four methods on most datasets. In contrast, the performance of STD fluctuates significantly on different scenes, especially in environments with more complex structures.

\subsection{Runtime Comparison with State-of-the-Arts}
\label{Runtime Comparison with State-of-the-Arts}
\par Table \ref{table3} provides the time consumption for descriptor extraction and loop detection, and then records the total runtime for processing each keyframe of IFTD and STD. It is evident that IFTD is approximately 50\% faster than STD in both descriptor extraction and loop detection. This significant speed improvement does not come at the expense of accuracy and robustness. These results highlight the dual advantages of IFTD in efficiency and effectiveness, demonstrating that our method not only processes data quickly but also provides reliable loop detection results while maintaining high precision and robustness.

\section{Conclusion}
\label{Conclusion}

\par In this work, we proposed a fast and robust Image Feature Triangle Descriptor (IFTD) for 3D point based loop detection in driving scenarios. By comparing with state-of-the-art loop detection methods on public datasets, we demonstrate that IFTD exhibits stronger robustness and significantly improved accuracy compared to these two methods. Additionally, our method achieves faster loop detection compared to STD. Future work will focus on dynamically adjusting feature extraction parameters based on scene information to achieve more robust feature extraction and descriptor construction. This improvement will further enhance the applicability and accuracy of our method, especially in handling scenes of different complexities and variations.

\bibliographystyle{IEEEtrans}
\bibliography{IEEEabrv,IEEEExample}




\ifCLASSOPTIONcaptionsoff
	\newpage
\fi



%

%


\begin{IEEEbiography}[{\includegraphics[width=1in,height=1.25in,clip,keepaspectratio]{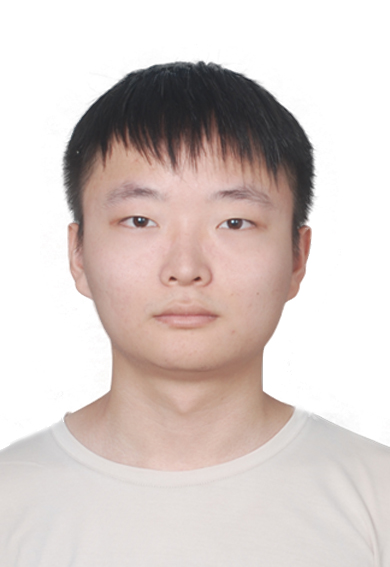}}]{Fengtian~Lang}
	received the B.E. degree from Huazhong University of Science and Technology (HUST), Wuhan, China, in 2023. He is currently a 1st year graduate student of HUST, School of Electronic Information and  Communications. He has published one paper on IROS and one paper on RAL. His research interests include LiDAR-inertial state estimation, LiDAR-inertial-wheel state estimation and LiDAR point based loop closing.
\end{IEEEbiography}


\begin{IEEEbiography}[{\includegraphics[width=1in,height=1.25in,clip,keepaspectratio]{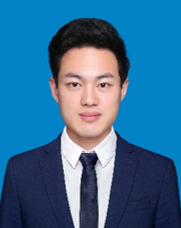}}]{Ruiye~Ming}
	received the B.E. degree from Zhengzhou University(ZZU), Zhengzhou, China, in 2022. He is currently a 2nd year graduate student of HUST, School of Electronic Information and  Communications. He has published one paper on RAL. His research interests include visual-LiDAR odometry, LiDAR point based loop closing, object reconstruction and deep-learning based depth estimation.
\end{IEEEbiography}


\begin{IEEEbiography}[{\includegraphics[width=1in,height=1.25in,clip,keepaspectratio]{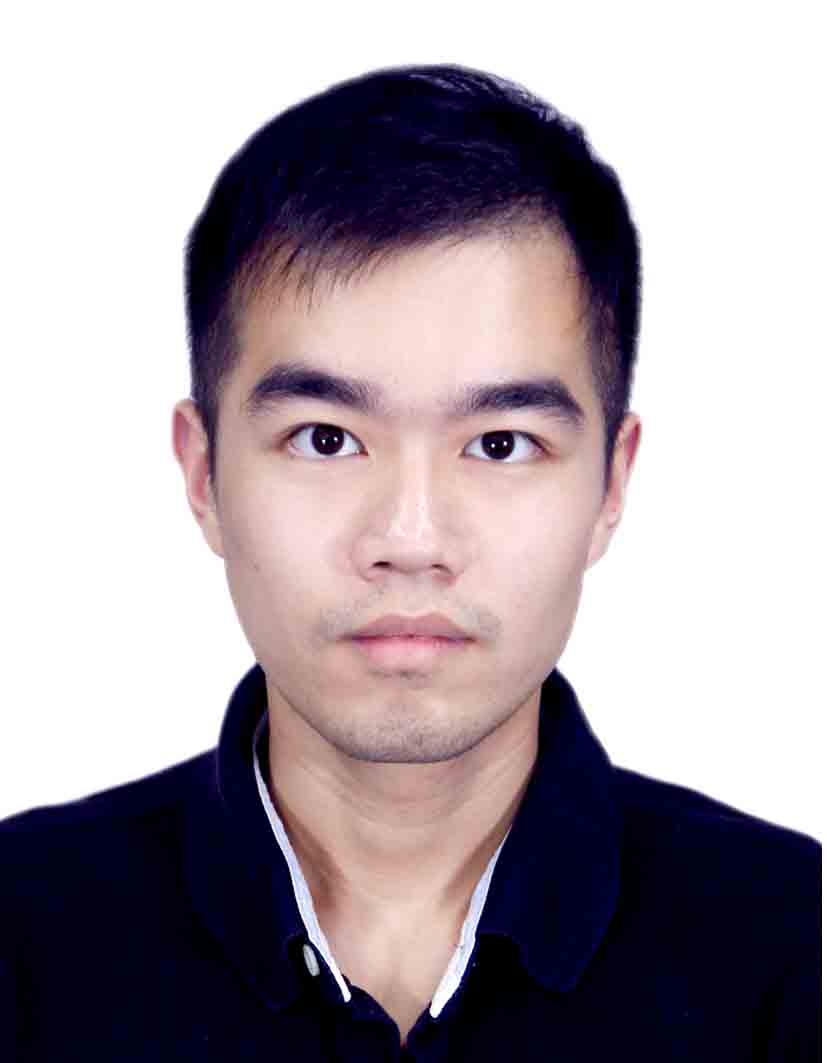}}]{Zikang~Yuan}
	received the B.E. degree from Huazhong University of Science and Technology (HUST), Wuhan, China, in 2018. He is currently a 4th year PhD student of HUST, School of Institute of Artificial Intelligence. He has published two papers on ACM MM, three papers on TMM, one paper on TPAMI, one paper on IROS and one paper on RAL. His research interests include monocular dense mapping, RGB-D simultaneous localization and mapping, visual-inertial state estimation, LiDAR-inertial state estimation, LiDAR-inertial-wheel state estimation and visual-LiDAR pose estimation and mapping.
\end{IEEEbiography}


\begin{IEEEbiography}[{\includegraphics[width=1in,height=1.25in,clip,keepaspectratio]{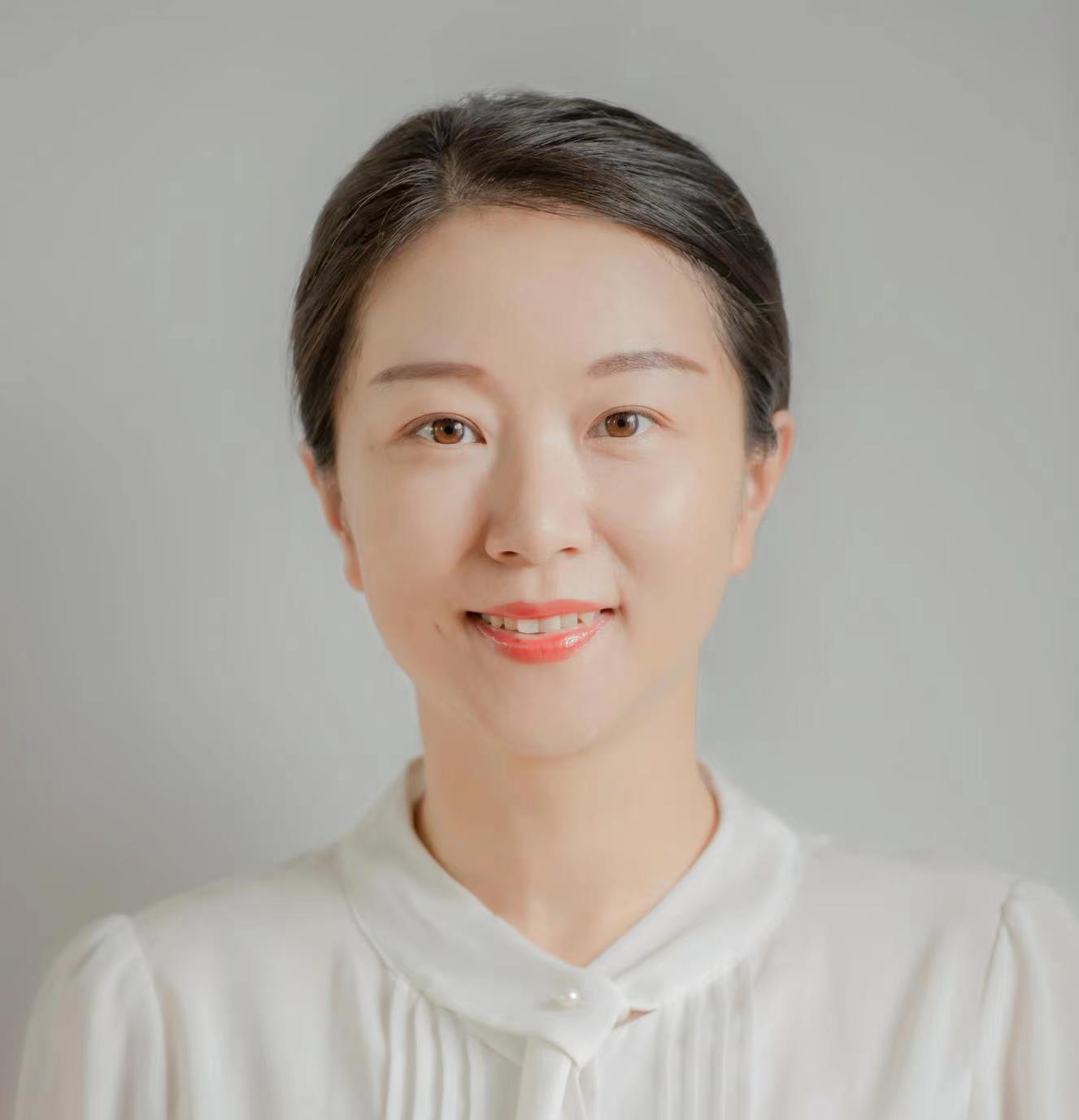}}]{Xin~Yang}
	received her PhD degree in University of California, Santa Barbara in 2013. She worked as a Post-doc in Learning-based Multimedia Lab at UCSB (2013-2014). She is current Professor of Huazhong University of Science and Technology School of Electronic Information and Communications. Her research interests include simultaneous localization and mapping, augmented reality, and medical image analysis. She has published over 90 technical papers, including TPAMI, IJCV, TMI, MedIA, CVPR, ECCV, MM, etc., co-authored two books and holds 3 U.S. Patents. Prof. Yang is a member of IEEE and a member of ACM.
\end{IEEEbiography}

\vfill



\end{document}